\documentclass{article}

\usepackage{arxiv}
\usepackage[numbers]{natbib}
\usepackage[utf8]{inputenc} % allow utf-8 input
\usepackage[T1]{fontenc}    % use 8-bit T1 fonts
\usepackage{hyperref}       % hyperlinks
\usepackage{url}            % simple URL typesetting
\usepackage{booktabs}       % professional-quality tables
\usepackage{amsfonts}       % blackboard math symbols
\usepackage{nicefrac}       % compact symbols for 1/2, etc.
\usepackage{microtype}      % microtypography
\usepackage{lipsum}		% Can be removed after putting your text content
\usepackage{graphicx}
\usepackage{natbib}
\usepackage{doi}
\newcommand\blfootnote[1]{%
  \begingroup
  \renewcommand\thefootnote{}\footnote{#1}%
  \addtocounter{footnote}{-1}%
  \endgroup
}
\usepackage{stackrel}
\usepackage{amsmath,amssymb,amsfonts}
\usepackage{textcomp}
\date{}

\title{Cram\'er-Rao bound-informed training of neural networks for quantitative MRI}

\author{Xiaoxia~Zhang$^{\dagger}$,
        Quentin Duchemin$^{\dagger}$,
        Kangning Liu$^{\dagger}$,
        Sebastian Flassbeck,
        Cem Gultekin, \\
        \AND 
        Carlos Fernandez-Granda,
        Jakob Assl\"ander
}

\renewcommand{\shorttitle}{Cram\'er-Rao bound-informed training of neural networks for quantitative MRI}

\begin{document}
\begin{center}
\textbf{Pre-Print - Submitted to Magnetic Resonance in Medicine}
\end{center}

\maketitle

\blfootnote{$\dagger$ Xiaoxia~Zhang, Quentin Duchemin, and Kangning Liu contributed equally to this work. \\
X. Zhang, S. Flassbeck, J. Assl\"ander, are with the NYU School of Medicine, Dept. of Radiology, NY, NYC. (e-mails: xiaoxia.zhang@nyulaonge.org, Sebastian.Flassbeck@nyulangone.org, Jakob.Asslaender@nyulangone.org)\\
Q. Duchemin, is with LAMA, Univ Gustave Eiffel, Univ Paris Est Creteil, CNRS, F-77447 Marne-la-Vallée,France.(e-mail:quentin.duchemin@univ-eiffel.fr)\\
K. Liu is with NYU Center for Data Science, NY, NYC. (e-mail:kangning.liu@nyu.edu)\\
C. Gultekin is with NYU Courant Institute of Mathematical Sciences, NY,NYC. (e-mail:cg3306@nyu.edu)\\
C. Fernandez-Granda is with NYU Center for Data Science and Courant Institute of Mathematical Sciences, NY,NYC.(e-mail:cfgranda@cims.nyu.edu)\\
This research was supported by the grant NIH/NIBIB R21 EB027241 and was performed under the rubric of the Center for Advanced Imaging Innovation and Research, a NIBIB Biomedical Technology Resource Center (NIH~P41 EB017183)
}

\begin{abstract}
Neural networks are increasingly used to estimate parameters in quantitative MRI, in particular in \textit{magnetic resonance fingerprinting}. Their advantages over the gold standard \textit{non-linear least square fitting} are their superior speed and their immunity to the non-convexity of many fitting problems. 
We find, however, that in heterogeneous parameter spaces, i.e. in spaces in which the variance of the estimated parameters varies considerably, good performance is hard to achieve and requires arduous tweaking of the loss function, hyper parameters, and the distribution of the training data in parameter space. 
Here, we address these issues with a theoretically well-founded loss function: the Cramér-Rao bound (CRB) provides a theoretical lower bound for the variance of an unbiased estimator and we propose to normalize the squared error with respective CRB. With this normalization, we balance the contributions of hard-to-estimate and not-so-hard-to-estimate parameters and areas in parameter space, and avoid a dominance of the former in the overall training loss. 
Further, the CRB-based loss function equals one for a maximally-efficient unbiased estimator, which we consider the ideal estimator. Hence, the proposed CRB-based loss function provides an absolute evaluation metric.
We compare a network trained with the CRB-based loss with a network trained with the commonly used means squared error loss and demonstrate the advantages of the former in numerical, phantom, and in vivo experiments. 
\end{abstract}

% keywords can be removed
\clearpage

\section{Introduction}
\label{sec:introduction}
% \IEEEPARstart{M}{agnetic} resonance fingerprinting (MRF) triggered a wave of research aimed at quantifying tissue properties in clinically feasible scan times
% \cite{ma2013magnetic}. 
% Key to an efficiency increase is to liberate the spin trajectories from the commonly used steady state and exponential decays and to allow for exploiting the entire Bloch sphere in search for an optimal encoding of the biophysical parameters of interest \cite{Asslander-hsCommPhysics,Asslander-hsMRM}. 
% Multiple recent studies have demonstrated this efficiency of MRF in quantifying tissue properties in various organs \cite{jiang2017repeatability,badve2015simultaneous,hamilton2017mr,chen2016mr}.

% The estimation of tissue parameters in MRF is a high-dimensional inverse problem where the parameters of interest, such as $T_1$ and $T_2$ relaxation times, are recovered from the corresponding fingerprint corrupted by k-space undersampling artefacts, thermal noise and field imperfection.
% This problem was originally tackled with template matching \cite{ma2013magnetic}. 
% Although dictionary matching ensures finding the global optimum within the simulated dictionaries, it has known practical challenges, such as discretization errors and increasing computation requirements with a growing number of biophysical parameters\cite{hamilton2019machine,song2019hydra}.

% (1)	qMRI quantifies parameter:
Quantitative MRI (qMRI) characterizes the spin physics in biological tissue with the aim to provide quantitative biomarkers for pathological changes \cite{barkhof2002clinico,tofts2005quantitative,matzat2013quantitative}. 
% (2)	the optimization landscape is usually non-convex:
qMRI entails fitting a biophysical model to a signal curve, and this model fitting is usually a non-convex problem, which is traditionally solved via non-linear least square (NLLS) fitting. Besides the risk of this iterative algorithm to get stuck in a local minimum, NLLS is often prohibitively slow for clinical routine imaging \cite{jelescu2016degeneracy}. 
Computation speed is particularly problematic for complex transient-state models that have been popularized by Magnetic Resonance Fingerprinting (MRF) \cite{ma2013magnetic}. 

% (4)	dictionary matching; or better NN
MRF outsources the slow simulations to the precomputation of a dictionary, which is then matched to the measured signal. This approach equals a brute force grid search, which also overcomes the issue of a non-convex optimization landscape.
Although dictionary matching ensures finding the global optimum within the simulated dictionary, it has known practical challenges, such as discretization errors and the so-called \textit{curse of dimensionality}. The latter describes the exponentially increasing memory and computation time requirements with a growing number of biophysical parameters\cite{hamilton2019machine,song2019hydra}.

The computational burden of dictionary matching can be reduced by singular value decomposition (SVD) and compressed sensing techniques \cite{cauley2015fast,yang2018low,davies2014compressed,mazor2018low,wang2016magnetic}. Nonetheless, these improvements do not overcome the curse of dimensionality and the computational burden is prohibitive for models with many parameters, such as the magnetization transfer (MT) model used here, which has overall 8 parameters. 
In contrast, the curse of dimensionality can be overcome with dictionary-free regression methods \cite{boux2018,nataraj2018}, as well as deep learning (DL). As a result, these methods are very fast once the network has been trained.
As the parameter fitting in both of these methods is feed-forward, it cannot get stuck in local minima. 
Because of those advantages over NLLS fitting and dictionary matching, these methods gained a lot of attention recently, in particular DL \cite{fang2019deep,fang2020submillimeter,hoppe2017,virtue2017better,cohen2018mr,hoppe2018deep,oksuz2019magnetic,hamilton2019machine,golbabaee2019geometry,song2019hydra,balsiger2018magnetic}:
e.g., the DRONE \cite{cohen2018mr} method maps the magnitudes of fingerprints to $T_1$ and $T_2$ maps with a four-layer fully connected network (FCN).
Virtue et al. \cite{virtue2017better} designed a three-layer FCN, which processes complex-valued data. 
During training, they augment the data with undersampling artifacts, which are heuristically derived from in vivo data. 
Additionally, convolutional neural networks \cite{hoppe2017deep,hoppe2018deep,hamilton2019machine} and recurrent neural networks \cite{oksuz2019magnetic} have been proposed to exploit the temporally local structure of the fingerprints. 
Inspired by the work of McGivney et al. \cite{mcgivney2014svd}, who showed that MRF dictionaries are often low rank, Golbabaee et al. \cite{golbabaee2019geometry} trained a neural network with a first layer fixed to the singular vectors associated with the highest singular values.
In \cite{song2019hydra}, image reconstruction is performed by solving a regularized version of an optimization problem with a low rank constraint, followed by a deep non-local residual convolutional neural network to restore parameter maps.

Most of the above described methods use the mean squared error (MSE) as the objective function during training, which aims to minimize the sum of the squared differences between each parameter and its estimate \cite{hoppe2017,song2019hydra,cohen2018mr,virtue2017better,oksuz2019magnetic,hoppe2017deep,hoppe2018deep,hamilton2019machine,golbabaee2019geometry}.
However, this loss has natural weaknesses for parameter estimation:
different parameters are at different scales and have different dimensions. For example, $T_1$ values are about 10 times larger than $T_2$. As a result, $T_1$ can dominate the MSE loss during the training of a network that jointly estimates both parameters \cite{fang2019deep,fang2020submillimeter}.
When estimating parameters of different physical dimensions, such as the fractional proton density $m_0^s$ (cf. Section \ref{sec:model}) and the relaxation times, using MSE as loss function becomes even more questionable from a physics perspective. Indeed, the use of MSE in this context violates the well-known homogeneity principle which states that two quantities with different dimensions cannot be added up.

These problems can be overcome with the mean relative absolute error as suggested in Refs. \cite{fang2019deep,fang2020submillimeter,gomez2020rapid}, or by training separate networks to estimate each parameter. Another problem is, however, not addressed by the mean relative absolute error: it is often easier to estimate one biophysical parameter within a specific range of parameters, than it is in other ranges. To give an example, the estimation of $T_1$ usually becomes increasingly difficult at short $T_2$-times as this reduces the overall signal to noise ratio. While this example is rather pictorial, the ease of parameter estimation is not always intuitive and gives rise to the same problem described above: the MSE is dominated by areas in parameter space where the estimate has large errors and these areas are usually not the ones of interest, in particular when using a pulse sequence that was optimized for a certain area in parameter space \cite{zhao2016optimal,Asslander-hsMRM}. 

In this paper, we introduce a theoretically grounded loss function that ensures close to optimal performance even in heterogeneous and high-dimensional parameter spaces. 
We will demonstrate that the proposed loss function fulfills these requirements by normalizing the squared error of each estimate with respective Cram\'er-Rao bound (CRB)\cite{rao1992information,cramerprinceton}, a theoretical lower bound for the variance of an unbiased estimator. 

In Section~\ref{sec:method} we introduce the CRB-weighted loss function while connecting neural-network (NN)-based parameter estimation back to signal processing theory. In Section~\ref{sec:experiment}, we show that our approach can jointly and efficiently predict multiple parameters in a high-dimensional parameter space, and we compare it to the commonly used MSE loss. We elaborate the advantages and disadvantages of the proposed loss function in Section ~\ref{sec:discussion}. Code for replicating the proposed work will be available on \href{https://github.com/quentin-duchemin/MRF-CRBLoss}{https://github.com/quentin-duchemin/MRF-CRBLoss}. 
% It was based on a MATLAB (Mathworks, USA) and Pytorch \cite{NEURIPS2019_9015} implementation. 
The most updated version of code for fingerprints simulation used in this paper is available on \href{https://github.com/JakobAsslaender/MRIgeneralizedBloch.jl}{https://github.com/JakobAsslaender/MRIgeneralizedBloch.jl}.

\section{Method}
\label{sec:method}

\subsection{The Cramér-Rao bound (CRB)}
\label{subsec:CRB}
First, we recap the definition of the Cram\'er-Rao Bound and some of its properties that are useful for this paper.
We consider a biophysical model with $P$ parameters and we denote the fingerprint corresponding to any set of tissue parameters $(\theta_1, \dots ,\theta_P) \in \mathbb R^P$ by $\mathbf x(\theta_1, \dots ,\theta_P) \in \mathbb C^d$, where $d$ is the number of data points in the fingerprint. We assume that for some tissue parameters $(\theta_1, \dots ,\theta_P)$ we observe a normally distributed random vector $X$ with mean $\mathbf x(\theta_1, \dots ,\theta_P) $ and with covariance matrix $\sigma^2 \mathrm{Id}_d$ where $\sigma^2>0$ and $\mathrm{Id}_d \in \mathbb R^{d\times d}$ is the identity matrix. We want to estimate $\theta_i$ (for some $i\in [P]$) from $X$. In general, an estimator of $\theta_i$ cannot minimize the MSE uniformly in $(\theta_1, \dots ,\theta_P)$, because of the bias-variance decomposition. However, if one restricts itself to the class of unbiased estimators, then the search for an estimator with minimal MSE
is reduced to the problem of variance minimisation. The CRB provides a universal limit for the noise variance of any unbiased estimator of the parameter $\theta_i$ \cite{rao1992information,cramerprinceton}.

We define the Fisher information matrix $F \in \mathbb{C}^{P\times P}$ at a point in parameter space $(\theta_1,\dots,\theta_P)$ whose entries are
\begin{equation}
    F_{i,j} :=  \frac{1}{\sigma^2}\left[\frac{\partial \mathbf x(\theta_1,\dots,\theta_P)}{\partial \theta_i}\right]^{H}\frac{\partial \mathbf x(\theta_1,\dots,\theta_P)}{\partial \theta_j}, \nonumber
\end{equation}
where  the superscript $H$ denotes the complex conjugate transposed. The Cramér-Rao bound associated with the $i^{\text{th}}$ parameter is defined as : $CRB_{i}(\theta_1,\dots,\theta_P) = \left(F^{-1} \right)_{i,i}.$
Given some $i\in [P]$, the noise variance of any unbiased estimator of $\theta_i$ based on the observation $X$ is at least as large as the corresponding Cramér-Rao bound $CRB_{i}(\theta_1,\dots,\theta_P)$.
% For any unbiased estimator, the estimate $\hat \theta_{i}$ of some parameter $\theta_{i}$ with $i \in [P]$ based on the observation $X$ (i.e. a fitting algorithm used to estimate $\hat{\theta}_{i}$ from $X$), the noise variance of $\hat \theta_{i}$ is at least as big as the corresponding Cramér-Rao bound.

\subsection{CRB-weighted MSE loss}
Ultimately, we aim at training a neural network that estimates parameters with high accuracy and precision. 
% The accuracy characterizes errors in estimates, measuring how close the mean of the estimated parameter is to the ground truth, and precision, which accounts for the spread, i.e., the standard deviation of the estimates.
High accuracy implies that the average of estimates over many noise realizations converges to the ground truth, i.e., the estimation has little-to-no bias. Precision analyzes the spread, i.e., the variance of estimates. From signal processing theory, we know that an unbiased estimator has a variance equal to or larger than the Cram\'er-Rao bound (CRB) (see Section~\ref{subsec:CRB}), and we have shown previously that the CRB is a good predictor of the noise variance for MRF-like data when using a non-linear least square fitting approach \cite{Asslander-hsCommPhysics}. 
We propose to normalize the squared error with respective CRB before averaging over all estimated parameters and all samples in the training data:
\begin{equation} \label{eq:crbloss}
L_{\text{CRB}} = \frac{1}{P_e S} \sum_{s=1}^S \sum_{p_e=1}^{P_e} \frac{(\hat{\theta}_{p_e,s} - \theta_{p_e,s})^2}{CRB_{p_e}(\theta_{1,s}, \ldots, \theta_{P,s})} .
\end{equation}
Here, $\theta$ denotes a biophysical parameter, $\hat{\theta}$ its estimate, 
$s \in \{1, \ldots, S\}$ runs over all samples in the training dataset,
$p_e \in \{1, \ldots, P_e\}$ over all parameters estimated by the network, and
$p \in \{1, \ldots, P\}$ over all parameters of the model.
The distinction between $P_e$ and $P$ is made to allow for estimating only a subset of parameters, which can be done while still considering a fit of the full model. 
The key here is to vary all parameters in the training dataset. In this case, the CRB has to account for the derivatives of the signal with respect to all model parameters. 

With this normalization, a maximally efficient unbiased estimator---which we consider the ideal estimator---has a loss of 1, which provides an absolute metric to evaluate a network's performance. 
Further it addresses above mentioned drawbacks of the MSE loss function: A maximally efficient unbiased estimator has a loss of 1 not only when averaging over all estimated parameters and all samples in the training dataset, but the expectation value of the CRB of each parameter and sample of the training dataset is one. Thus, the CRB-weighted loss function suffers neither from being dominated by parameters with large values, nor by parameters that are difficult to estimate.

\subsection{Biophysical model} \label{sec:model}
In order to highlight the ability of our loss function to handle high-dimensional parameter spaces in which the difficulty to estimate parameters varies substantially, we use an 8-parameter magnetization transfer model \cite{Asslander2019ismrm}. It is based on Henkelman's original two-pool spin model\cite{henkelman1993quantitative} that distinguishes between protons bound in water---the so-called \textit{free pool}---and protons bound in macromolecules, such as proteins or lipids---the so-called \textit{semi-solid pool}.
The pulse sequence is designed such that the free pool remains in the hybrid state \cite{Asslander-hsCommPhysics, Asslander2019ismrm}---a spin ensemble state that provides a combination of robust and tractable spin dynamics with the ability to encode biophysical parameters with high signal-to-noise ratio (SNR) efficiency compared to steady-state MR experiments \cite{Asslander-hsCommPhysics}. 
The model has the following parameters: An apparent $T_1$ relaxation time of both pools, $T_2^f$ of the free and $T_2^s$ of the semi-solid pool, the size of the semi-solid pool $m_0^s$, which is normalized by the sum $m_0^s + m_0^f = 1$, the exchange rate $R_\text{x}$ between the two pools, the imperfectly calibrated $B_0$ and $B_1$, and a complex-valued scaling factor $M_0$. 

\subsection{Data simulation}
We simulated fingerprints with a custom implementation in MATLAB (Mathworks, USA) with random sets of parameters with the following distributions: we used truncated Gaussian distributions with means and standard deviations of $m_0^s = 0.12\pm0.08$ while ensuring $m_0^s \geq 0$, $T_1 = (1.6\pm0.8)$s, $T_2^f = (0.1\pm0.2)$s while ensuring $T_2^f \leq T_1$, $R_\text{x} = (44\pm20)$/s while ensuring $R_\text{x} \geq 0$, $B_1/B_1^{\text{nominal}} = 1\pm0.3$. Further, we used a uniformly distributed $B_0 \in [-2\pi/T_\text{R}, 2\pi/T_\text{R}]$ where $T_\text{R}$ is the repetition time.

With these distributions, we simulated 92,160 fingerprints for a training dataset, 10,240 for a validation, and 9,056 for our testing dataset \#1. 
We performed a singular value decomposition of the training dataset and compressed all three datasets to the coefficients corresponding to the 13 largest singular values. 

After computing the SVD, we multiplied the three datasets with a random scaling factor $M_0$, which has a uniformly distributed absolute value $|M_0| \in [0.1, 1]$ and a complex phase uniformly distributed in the range $[0, 2\pi]$. We added complex valued Gaussian noise with a standard deviation of 0.01, which results in an overall SNR\textsubscript{max} in the range 10 to 100, where we define SNR\textsubscript{max} as the maximum achievable SNR, i.e. the SNR one would measure with $T_\text{R} \rightarrow +\infty$ and the echo time $T_\text{E} \rightarrow +0$ \cite{Asslander2020}. Note that we multiplied the fingerprints with a different $M_0$ and we added different noise realizations in each training epoch to reduce overfitting. 

In addition to the randomly distributed testing dataset \#1, we also conducted analyses on a regular grid for a simple visualization of certain performance metrics. This dataset \#2 is limited to 2D slices that cut through the 8-dimensional parameter space: one slice varies $m_0^s$ and $T_1$ and one varies $T_1$ and $T_2^f$ while fixing other 6 parameters to the mean values used in the training sampling scheme.

\subsection{Neural network design}
\label{sec:network design}
As our quantitative MT model is more complex compared to the Bloch model used in previous NN-based MRF\cite{hoppe2017,cohen2018mr,golbabaee2019geometry,virtue2017better,song2019hydra}, we use a larger network with more capacity to capture the high-dimensional mapping functions. The network size was empirically selected to ensure accurate functional mapping while keeping the training time and memory requirements within limits.
% In the following, we will refer to this network as \textit{golden network}. 
Our network consists of 14 fully-connected layers. 
%(e.g., we modified the network to output B0 and B1 for creating Fig. XX).
Except for the output layer, each fully-connected layer is followed by group normalization \cite{wu2018group} and rectifier linear units (ReLU)  activation functions \cite{agarap2018deep}. We incorporated skip connections \cite{he2016deep} to avoid the vanishing gradient problem during training. 

The network used here treats each voxel independently. The inputs of the network are the 13 complex-valued coefficients of the compressed training or testing data, split into real and imaginary parts and normalized to have an $\ell_2$-norm of 1. 
The outputs of the last layer are the estimated parameters of interest; in our case $m_0^s$, $T_1$, and $T_2^f$ as we consider these parameters to be the most relevant to our main target application multiple sclerosis and since we optimized the pulse sequence for this purpose \cite{Asslander2019ismrm}. The network is, however, also capable of estimating additional parameters, such as $B_0$ and $B_1$, with the same training routine and architecture, modified to have more output channels (not shown here).
% 
% \begin{figure}[!h]
% \centering
% \captionsetup{justification=centering}
% \vspace{-1em}
% \includegraphics[scale=0.50]{NNstruct.png}
% \caption{Network structure. } \label{NNstruct}
% \end{figure} 

\subsection{Training details}
The weights of the network are initialized randomly and we used ADAM optimizer\cite{kingma2014adam} with a batch size of 512. The learning rate was initially set to 0.01 and decayed by half every 40 epochs. We trained the network for 400 epochs, and observed good convergence. 
We trained two networks: one with proposed CRB-weighted loss (see~\eqref{eq:crbloss}) and one with the commonly used MSE loss for comparison. We experimentally found this network is not very sensitive to hyper-parameters by searching the hyper-parameter space, therefore, we set them identical when train both networks.
% We used identical hyper-parameters for training both networks. 

\subsection{Bias and variance analysis}
In order to separate bias from noise in our performance analysis, we process each fingerprint of the two testing datasets with $N$ different noise realizations for a given SNR\textsubscript{max}. This allows us to calculate the bias of a parameter estimate $\hat{\theta}_{p_e,s}$:
\begin{equation}
\text{bias}(\theta_{p_e,s}) = \bar{\theta}_{p_e,s} -\theta_{p_e,s},
\label{eq:bias}
\end{equation}
as well as the variance
\begin{equation}
\sigma^2(\theta_{p_e,s}) =  \sum_{n=1}^{N} \frac{ (\hat{\theta}_{n,p_e,s}-\bar{\theta}_{p_e,s})^2}{N}, 
\label{eq:variance}
\end{equation}
where 
$n \in \{1, \ldots, N\}$ runs over all noise realizations,
$\theta_{p_e,s}$ denotes the ground truth,
$\hat{\theta}_{n,p_e,s}$ an estimate, and
$\bar{\theta}_{p_e,s}$ the average of all $N$ estimates of $\theta_{p_e,s}$. 

In the same spirit, multiple noise realizations of a single fingerprint allow us to split the average loss of each fingerprint into a bias and a variance component:

\begin{align} \label{eq:biasvariance}
\bar{L}_{\text{CRB}}(\theta_{p_e,s}) = \frac {\sum_{n=1}^{N}(\hat{\theta}_{n,p_e,s} - \theta_{p_e,s})^2}{N \cdot CRB_{p_e}(\theta_{1,s},\dots,\theta_{P,s})}\stackrel{N \rightarrow +\infty}{=} \underbrace{\frac{(\bar{\theta}_{p_e,s}-\theta_{p_e,s})^2}{CRB_{p_e}(\theta_{1,s},\dots,\theta_{P,s})}}_{\bar{L}_{\text{CRB}}^\text{bias}(\theta_{p_e,s})} +\underbrace{\frac{\sum_{n=1}^{N}(\hat{\theta}_{n,p_e,s}-\bar{\theta}_{p_e,s})^2}{N \cdot CRB_{p_e}(\theta_{1,s},\dots,\theta_{P,s})}}_{\bar{L}_{\text{CRB}}^{\sigma^2}(\theta_{p_e,s})}
\end{align}

\noindent where $\bar{L}_{\text{CRB}}^\text{bias}(\theta_{p_e,s})$ and $\bar{L}_{\text{CRB}}^{\sigma^2}(\theta_{p_e,s})$ are the contributions of the bias and the variance to the average loss of each parameter and sample $\theta_{p_e,s}$, respectively.
The bar indicates the loss averaged over multiple noise realizations.

Averaging the loss further over all estimated parameters $p_e$ and samples or fingerprints $s$ results in the overall loss with its two contributions:
\begin{equation} \label{eq:biasvariance2}
\bar{L}_{\text{CRB}} =  \underbrace{\sum_{p_e,s} \frac{\bar{L}_{\text{CRB}}^{\text{bias}}(\theta_{p_e,s})}{P_e \cdot S}}_{\bar{L}_{\text{CRB}}^\text{bias}} + \underbrace{\sum_{p_e,s} \frac{\bar{L}_{\text{CRB}}^{\sigma^2}(\theta_{p_e,s})}{P_e \cdot S}}_{\bar{L}_{\text{CRB}}^{\sigma^2}} .
\end{equation}
Note that $\bar{L}_{\text{CRB}} = \sum_n L_{\text{CRB}, n} /N$ connects this average loss back to the one used during training (Eq.~\eqref{eq:crbloss}). 

% We used Eq. \eqref{eq:bias_variance2} to calculate the contributions of bias and variance to the loss for different SNR\textsubscript{max} values by evaluating $N=300$ noise realizations for each noise level and we repeated the analysis with the MSE-loss using Eqs. \eqref{eq:bias_variance} and \eqref{eq:bias_variance2} without weighting each component by CRB values. 
% For SNR\textsubscript{max} = 50, we further analyzed the bias and variance contributions of each parameter to the loss.

We used Eq. \eqref{eq:biasvariance2} to calculate the contributions of bias and variance to the CRB-loss for different SNR\textsubscript{max} values by evaluating $N=300$ noise realizations for each SNR\textsubscript{max}.
The same analysis was repeated for the MSE-loss without normalizing the CRB values in Eqs. \eqref{eq:biasvariance}. 
To further investigate those contributions of each parameter to the loss, we conducted this analysis based on each parameter separately for a specific SNR\textsubscript{max}.

\subsection{Phantom and in vivo scans}
We built a magnetization transfer phantom with thermally cross-linked bovine serum albumin (BSA). 
We mixed BSA powder with distilled water in three different concentrations: 
10\%, 15\%, and 20\% of the overall sample weight.
The mixtures were stirred at 30$^\circ$C until the BSA was fully dissolved. 
We split each solution into two batches and doped one of them with 0.1mM MnCl\textsubscript{2}.
We filled six plastic tubes with the resulting solutions and thermally cross-linked them in a water bath at approximately 90$^\circ$C for 10 minutes.
We note that the 10\% BSA mixture without MnCl\textsubscript{2} was cross-linked separately as a trial run, which seemed to have resulted in somewhat inconsistent MT properties. 
We immersed the six tubes in a head-sized cylindrical container filled with doped water. 

We scanned the phantom with our hybrid-state qMT sequence \cite{Asslander2019ismrm} on a 3T Prisma scanner (Siemens Healthineers, Erlangen, Germany) with a 20-channel head coil. The sequence acquires 3D data with 1mm isotropic resolution in approximately 12 minutes with a radial koosh-ball k-space trajectory, whose angles are incremented by 2D-golden angles \cite{Asslander2019ismrm,Sebastian2021ISMRM, Winkelmann2007, Chan2009}.

We further scanned an asymptomatic volunteer with approval of our institutional review board and after getting informed consent. For the in vivo scan, we used a 64-channel head-neck coil and we compressed the data to 20 virtual coils with a singular value decomposition.

We reconstructed both datasets with the \textit{low rank inversion} described in Ref. \cite{asslander2018low}, using the 13 singular vectors from above described SVD of the training data. 
We used the \textit{BART} implementation of this reconstruction \cite{tamir2017t2,ueckerbart} and added locally-low rank regularization to reduce undersampling artifacts and noise \cite{tamir2017t2}.
The 13 resulting coefficient images show minimal undersampling artifacts and we assume the residual artifacts and noise to be Gaussian distributed.
Voxel-by-voxel, these 13 complex-valued coefficients, normalized to have an $\ell_2$-norm of 1, are fed into the neural networks for the parameter estimation. 

To analyze the phantom data, we selected a central slice through the phantom and masked each tube separately. We eroded the out-most voxels to avoid partial-volume effects and performed a box-plot analysis. Thereafter, we compare parameters estimated with a non-linear least square fit (NLLS), with the neural network that was trained with the MSE loss, and with the neural network trained with the CRB-weighted loss.

%%%%%%%%%%%%%%%%%%% Experiments and results %%%%%%%%%%%%%%%%%%%%%%%%
\section{Results}
\label{sec:experiment}
%%%%%%%%%%%%%%%%%%%%%%%%%%%%Comparison with MSE %%%%%%%%%%%%%%%%%%%%%%%%%%%%%%%%%%%%
\subsection{Convergence analysis}
\begin{figure}[htbp]
\centering
\vspace{-1em}
\includegraphics[scale=0.45]{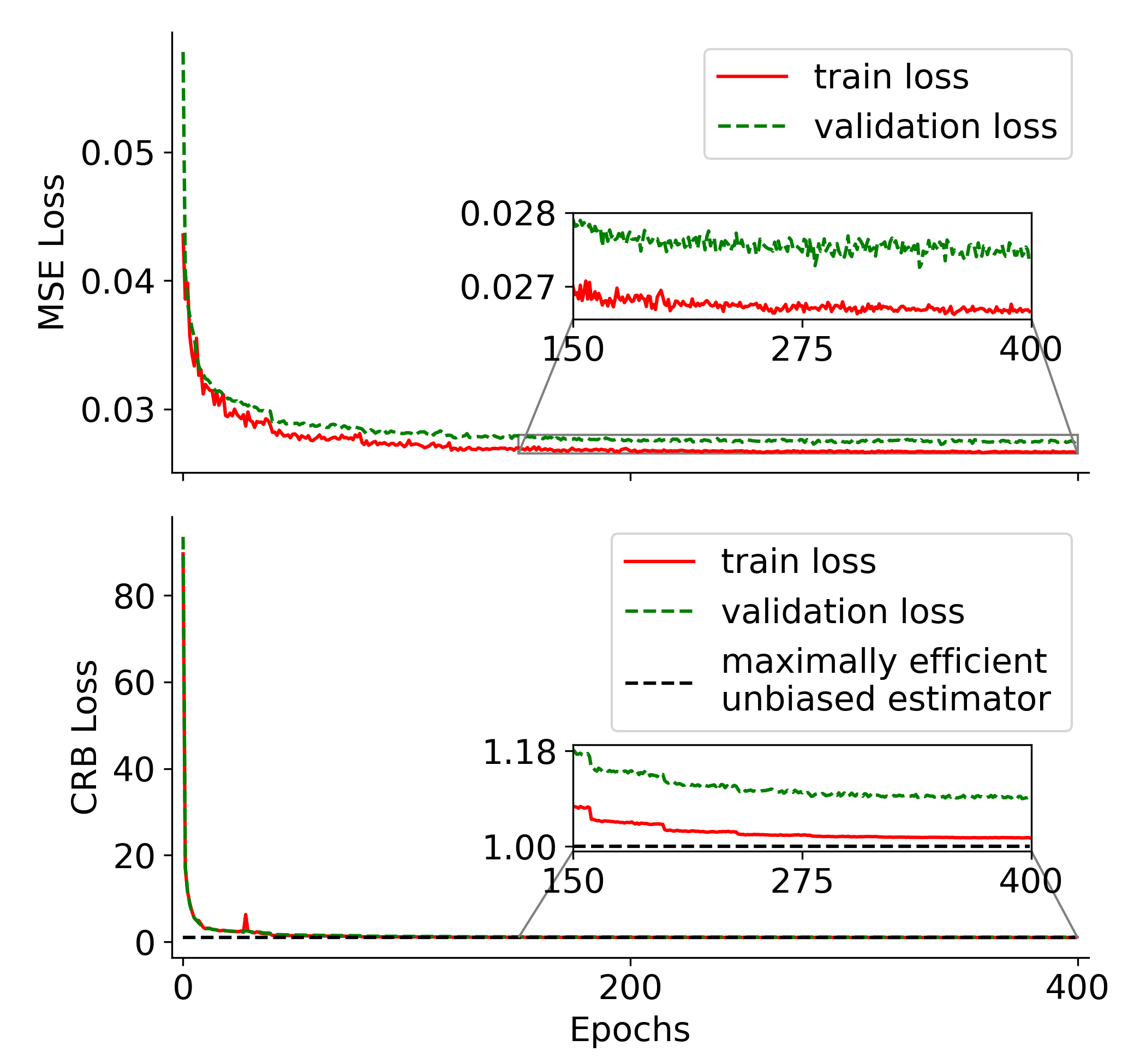}
\caption{Convergence of the training and validation loss. The CRB-loss converges to approximately 1, which corresponds to the loss of a maximally efficient unbiased estimator, while the MSE-loss converges to a value that provides little insight in the performance of the network. 
The two curves result from separate networks trained with respective loss function. 
} 
\label{fig:loss}
\end{figure}

Fig. \ref{fig:loss} reveals that the CRB-weighted loss indeed converges approximately to 1, which is a necessary (but not sufficient) condition of a maximally efficient unbiased estimator and, thus, provides an absolute evaluation metric. In contrast, the MSE-loss converges to a value that gives little insight in the performance of the network. Additionally, the CRB-loss converges virtually monotonously while the MSE-loss exhibits comparably strong fluctuations. 

% %%%%%%%%%%%%%%
\subsection{Bias and variance analysis of the converged networks}
\begin{figure}[htbp]
\centering
\vspace{-1em}
\includegraphics[scale=0.4]{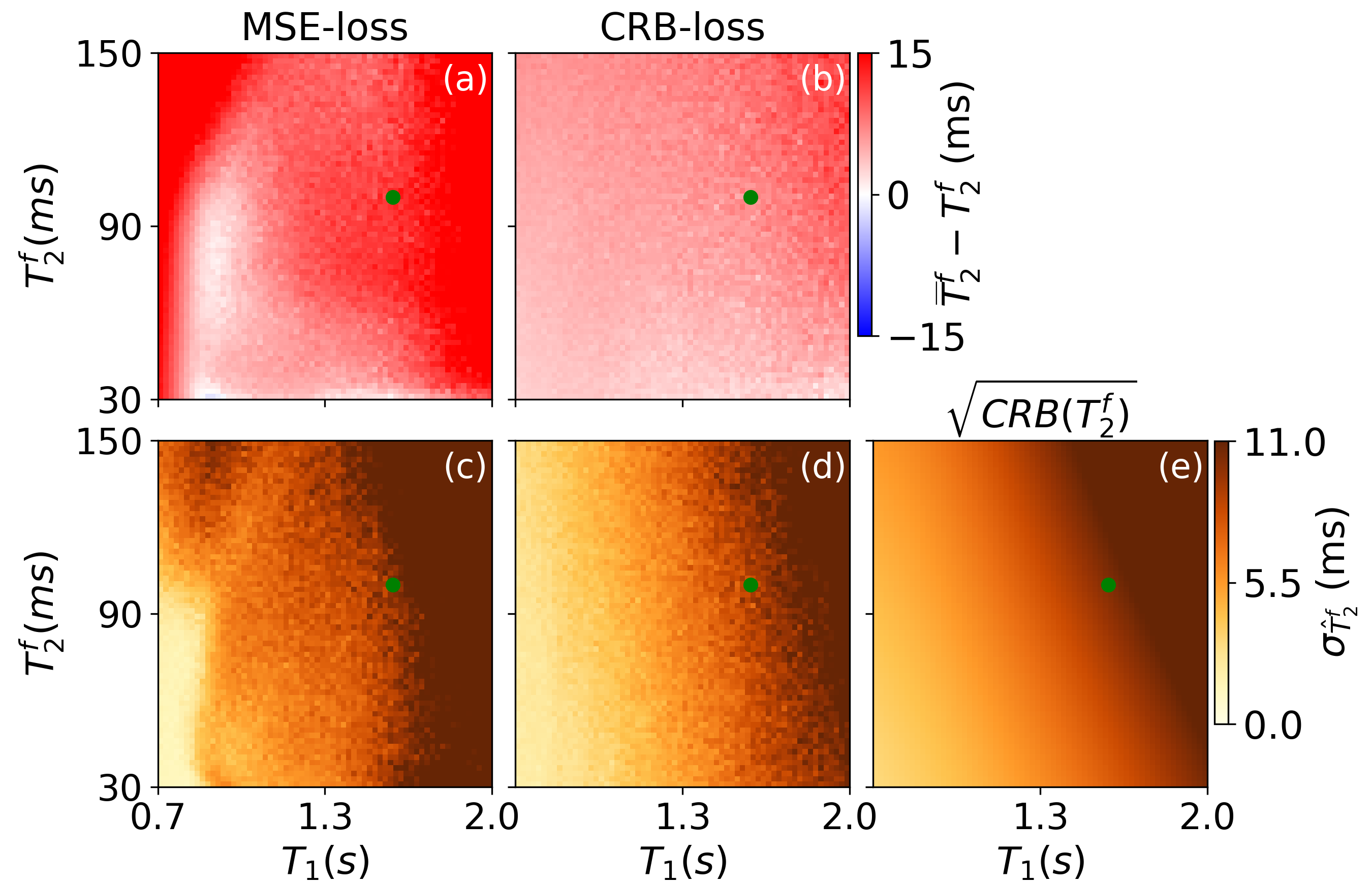}
\caption{Bias (a,b) and standard deviation (c,d) of $\hat{T}_2^f$, estimated with the networks trained with the MSE-loss and CRB-loss, respectively. The standard deviation is compared to the square root of the Cram\'er-Rao bound (e), which provides a theoretical lower bound for an unbiased estimator. The green dots indicate the mean values of the corresponded parameters in the training dataset.
The maps were generated with the test dataset \#2.} \label{fig:T2heatmap}
\end{figure}

In order to confirm that the network trained with the CRB-loss approximates a maximally efficient unbiased estimator, we performed three analyses. 
The first one aims at a visual analysis and uses the testing dataset \#2 that lies on a regular grid. As the parameter space is 8-dimensional, this analysis is limited to single slices through this space. 
When using the network trained with the CRB-loss, the bias of the $T_2^f$ estimation in a slice spanned by $T_1$ and $T_2^f$ is small (Fig.~\ref{fig:T2heatmap}b; 5.23ms on average) compared to the bias of the estimation with the MSE-based network (Fig.~\ref{fig:T2heatmap}a; 14.1ms on average).

%heatmap T2

Comparing the standard deviation of estimates to the square root of the Cram\'er-Rao bound, we observe close concordance in the case of the network trained with the CRB-loss, while we observe substantial deviations for the network trained with the MSE-loss (Fig.~\ref{fig:T2heatmap}c-e). In particular at short $T_1$ and long $T_2^f$ times, the standard deviation of $\hat{T}_2^f$, estimated with the MSE-based network, is substantially larger compared to the square root of the CRB, indicating sub-optimal precision in addition to the large bias (Fig.~\ref{fig:T2heatmap}a,c). In contrast, when using the network trained with the CRB loss, we find good agreement between standard deviation of the parameter estimates and the square root of the CRB itself, which indicates that this network approximates a maximally efficient unbiased estimator.
Overall, the two networks have similar performance in estimating $m_0^s$ and $T_1$ on the dataset \#2 (cf. supplementary material), while we do observe a substantial difference in the performance in estimating $T_2^f$.

\begin{figure}[tbp]
\centering
\vspace{-1em}
\includegraphics[scale=0.44]{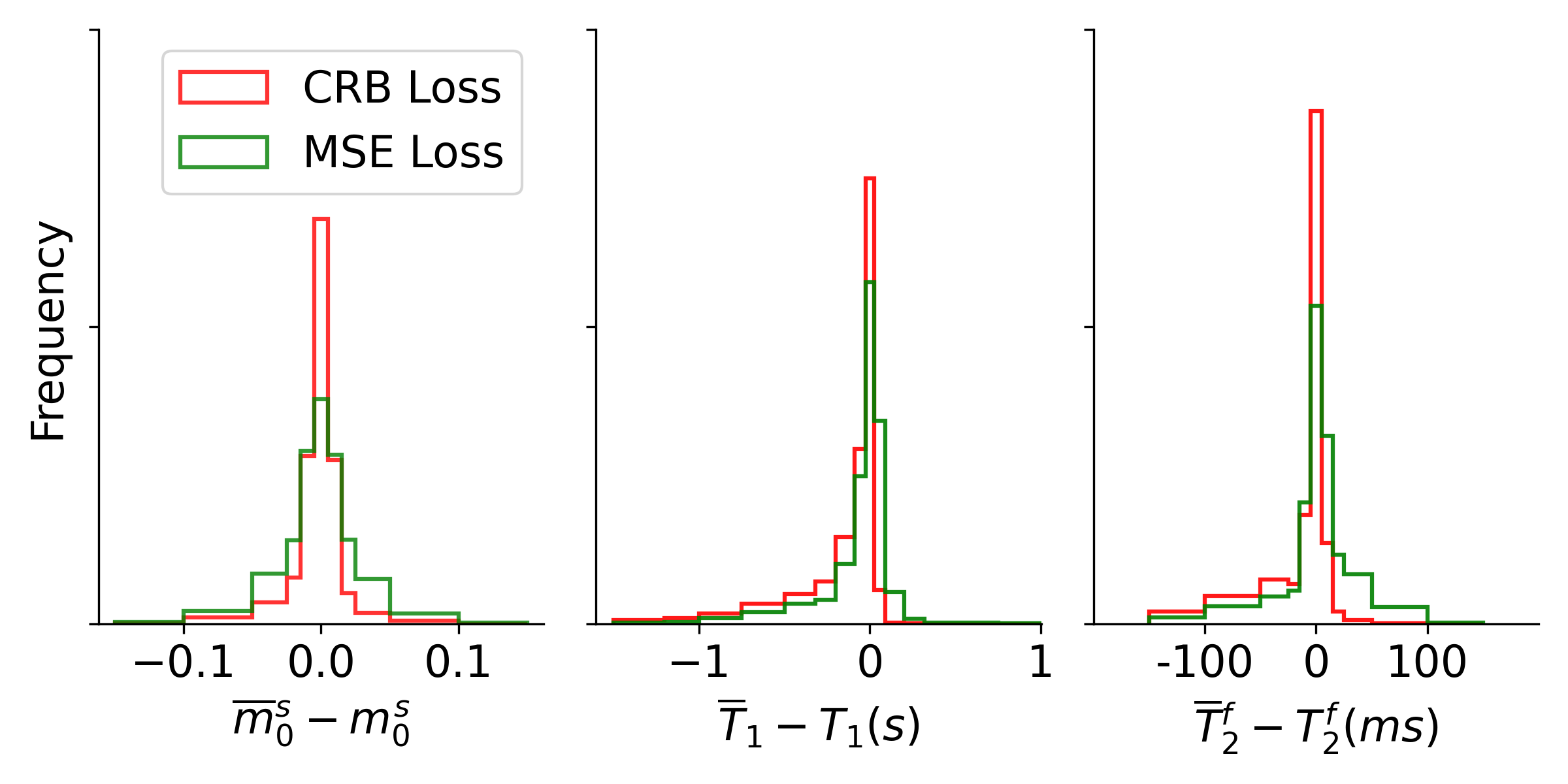}
\caption{Bias analysis. The randomly sampled fingerprints in test dataset \#1 were processed with 300 noise realizations (SNR\textsubscript{max}$=50$) and the mean value is compared to the ground truth. Overall, one can observe a smaller bias when estimating the parameters with the network trained with the CRB-loss compared to the network trained with the MSE-loss.} \label{fig:validation_bias}
\end{figure}

These findings are confirmed by an analysis of the test dataset \#1, which covers the same volume in the 8D parameter space as the training data. The bias of $m_0^s$, visualized with histograms in Fig.~\ref{fig:validation_bias}, is overall smaller when using the CRB-based network in comparison to the MSE-based network. 
The same holds true for $T_1$ and $T_2^f$, as evident by the higher count in the bin centered around zero bias. This finding is, however, somewhat obscured by the opposite signs in the biases when comparing the two networks.

\begin{figure}[tbp]
\centering
\vspace{-1em}
\includegraphics[scale=0.44]{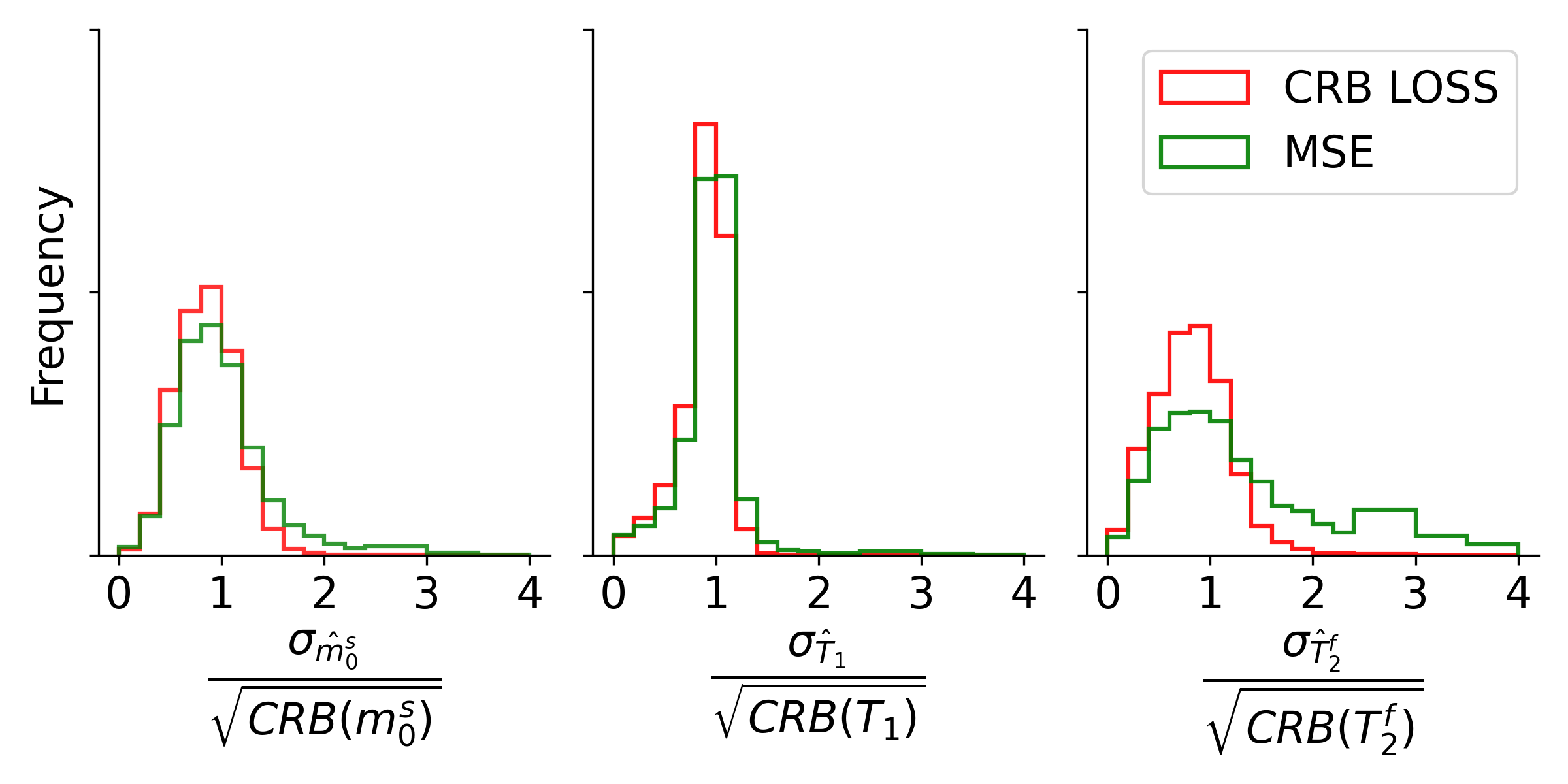}
\caption{Standard deviation analysis. The randomly sampled fingerprints in test dataset \#1 were processed with 300 noise realizations (SNR\textsubscript{max}$=50$) and the standard deviation of the estimates is analyzed. Overall, one can observe a smaller variance when estimating the parameters with the CRB-based network compared to the network trained with the MSE-loss. A maximally efficient unbiased estimator has a standard deviation, divided by the square root of the CRB, of 1.} \label{fig:validation_variance}
\end{figure}

% \begin{table}
% \caption{Estimates error of the two networks for the simulated datasets}
% \label{tab1}
% % \setlength{\tabcolsep}{3pt}
% \begin{tabular}{p{30pt}|p{50pt}|p{55pt}|p{55pt}}
% \hline
% dataset &paramter     &MSE-loss    &CRB-loss \\
% \hline
% \multirow{3}{*}{\makecell{Testing\\dataset \#1}} 
% &$m_0^s$       &0.10$\pm$0.071     &0.11$\pm$0.072 \\
% &$T_1(s)$      &1.71$\pm$0.45      &1.79$\pm$0.49  \\
% &$T_2^f(ms)$    &105.0$\pm$32.2     &99.4$\pm$45.3 \\
% \hline
% \hline
% \multirow{3}{*}{\makecell{Testing\\dataset \#2}} 
% &$m_0^s$      &0.013$\pm$0.005      &0.0053$\pm$0.0031 \\
% &$T_1(s)$     &0.069$\pm$0.066      &0.113$\pm$0.12  \\
% &$T_2^f(ms)$   &19.3$\pm$7.2         &5.2$\pm$2.7 \\
% \hline
% \multicolumn{4}{p{240pt}}{Testing dataset \#1 is randomly distributed in the 8-dimensional parameter space and testing dataset \#2 are 2D slices uniformly sampled in the space. }\\
% \multicolumn{4}{p{240pt}}{Data are summarized as mean $\pm$ standard deviation of root square error.}\\
% \end{tabular}
% \label{tab1}
% \end{table}

Fig. \ref{fig:validation_variance} depicts the ratio of the estimates' standard deviation and the square root of the CRB. For a maximally efficient unbiased estimator, this ratio is 1 and the network trained with the CRB-loss approximates this property well and better compared to the network trained with the MSE-loss, in particular in the estimation of $T_2^f$.

\begin{figure}[tbp]
\centering
\vspace{-1em}
\includegraphics[scale=0.55]{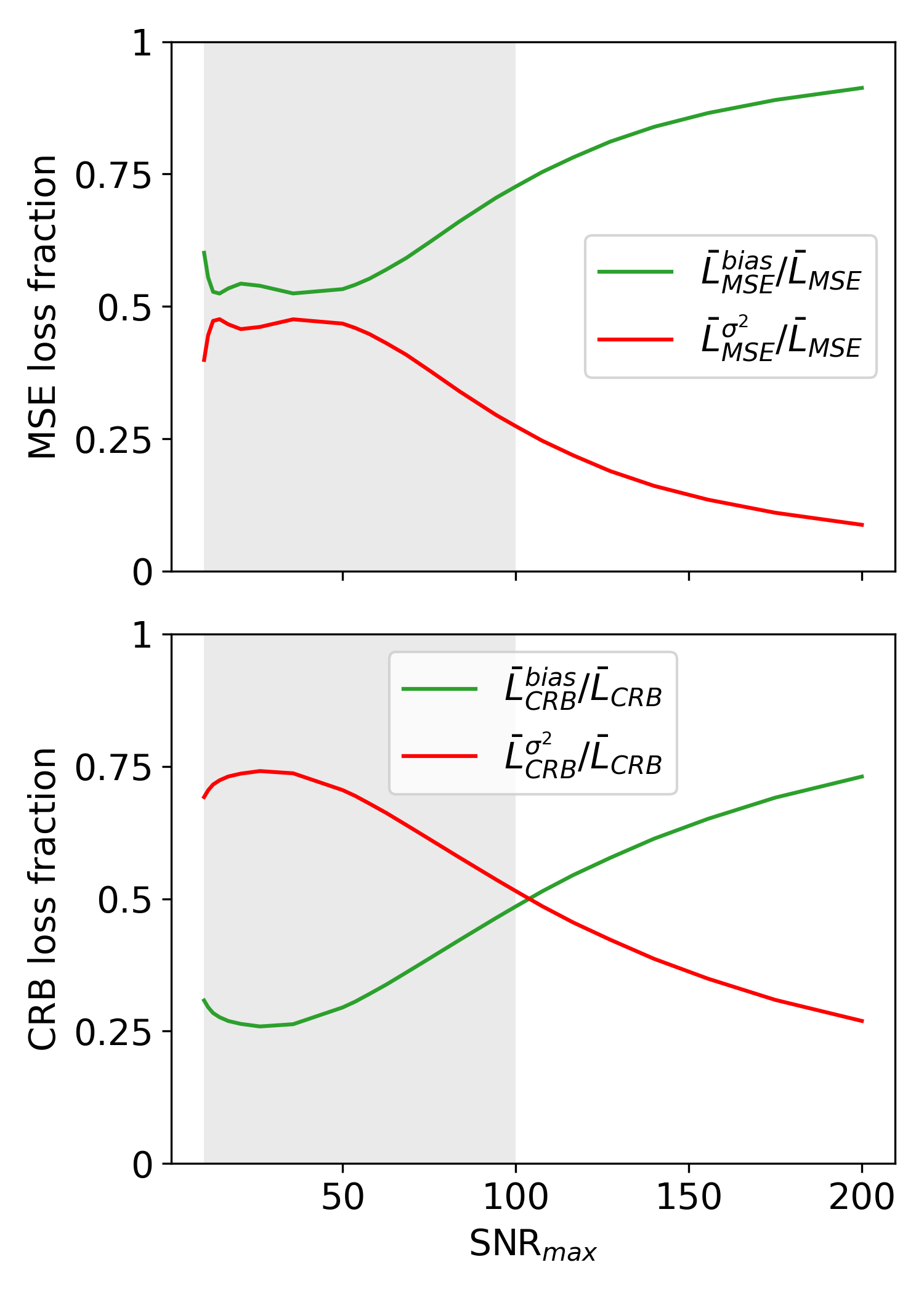}
\caption{Bias and variance contributions to the loss.
In case of the network trained with the MSE-loss, the bias dominates the overall loss. 
In contrast, the loss of the CRB-based network is dominated by the variance within the training range SNR\textsubscript{max} $\in [10, 100]$, highlighted by the gray shade. 
This decomposition of CRB-loss was performed with Eq.~\eqref{eq:biasvariance2} on the test dataset \#1. 
} \label{fig:noise_level}
\end{figure}

In the third analysis, we decomposed the loss with Eq.~\eqref{eq:biasvariance2} into a bias and a variance component. In the case of the MSE-based network, the loss is dominated by the bias. 
In contrast, the loss of the CRB-based network is dominated by the variance within the training range of SNR\textsubscript{max} $\in [10, 100]$ and the bias component becomes dominant only at SNR\textsubscript{max} values outside of the training range (Fig. \ref{fig:noise_level}). 

\begin{figure}[tbp]
\centering
\vspace{-1em}
\includegraphics[scale=0.47]{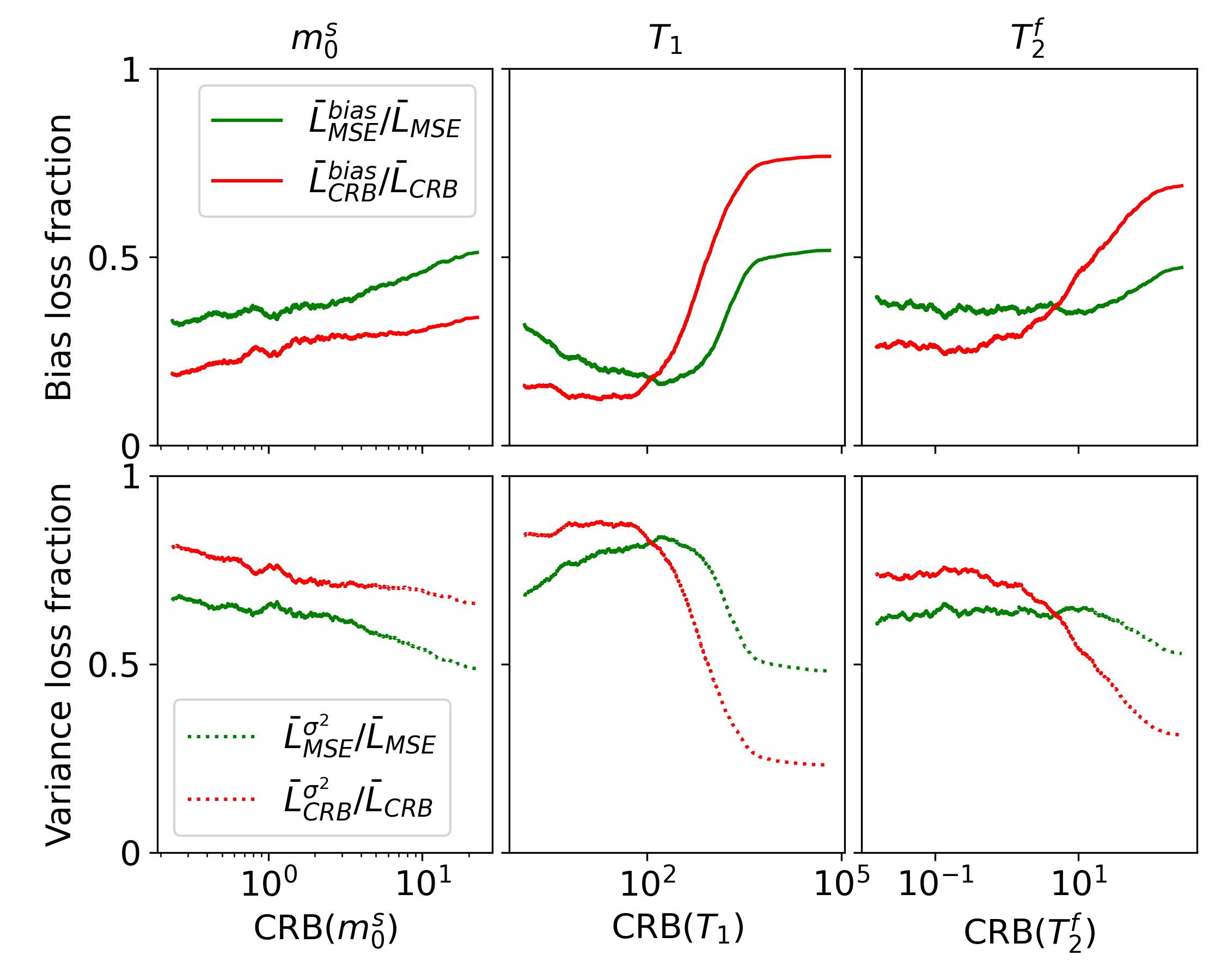}
\caption{Bias and variance contributions to the loss. 
The CRB-based network results overall in a smaller bias compared to the MSE-based network, with the exception of $T_1$ and $T_2^f$ at very high CRB-values, i.e. for parameter combinations that are hard to estimate. 
The decompositions of the loss were performed with Eqs.~\eqref{eq:biasvariance} and \eqref{eq:biasvariance2} on the test dataset \#1 with SNR\textsubscript{max} = 50 and is further split into separate parameters.
} \label{fig:crb_level}
\end{figure}

Taking a close look at the loss composition at SNR\textsubscript{max} = 50, which is roughly the SNR found in vivo, we find that the bias contributions are overall lower for the CRB-based network, with the exception of $T_1$ and $T_2^f$ at very high CRB values (Fig. \ref{fig:crb_level}). This confirms that, at least for parameter combinations with moderate CRB values, the CRB-based network results on average in a smaller bias. 

\subsection{Phantom data}
\begin{figure}[htbp]
\centering
\vspace{-1em}
\includegraphics[width=1.0\linewidth]{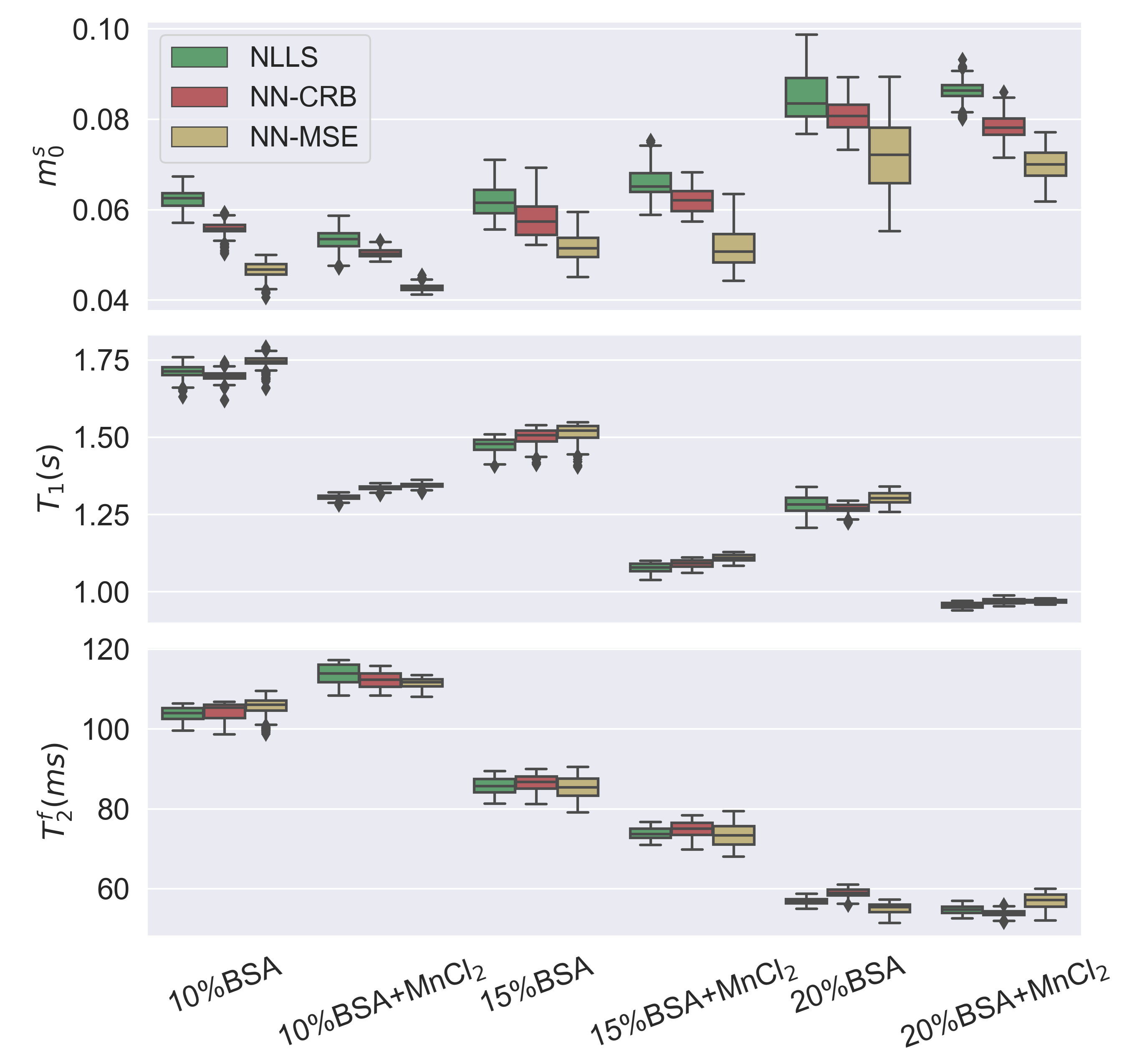}
\caption{Estimates of $m_0^s$, $T_1$, and $T_2^f$ from a custom phantom containing different concentrations of thermally cross-linked bovine serum albumin (BSA), half of them doped with MnCl\textsubscript{2}.
The three methods analyzed here show overall good agreement, but the neural network (NN) trained with the CRB loss is consistently in better agreement with the non-linear least square (NLLS) fits, which we consider the gold standard.
} \label{fig:phantom}
\end{figure}

The improved performance of the network trained with the CRB-loss is confirmed by phantom experiments (Fig.~\ref{fig:phantom}). 
We estimated $m_0^s$, $T_1$, and $T_2^f$ for the samples with different BSA concentrations using the two neural networks, as well as a non-linear least square (NLLS) fitting algorithm, which we consider the gold standard.  
For virtually all samples, we found better agreement of the CRB-based network estimates with the NLLS results, compared to the MSE-based network estimates. 
For most samples, the estimates with the CRB-based network and the NLLS algorithm match within their interquartile range. 
The most pronounced deviations can be found in $m_0^s$, which has overall the highest CRB and is, thus, the most difficult one to estimate (not shown here). 
Estimates calculated with the MSE-based network are overall in good agreement with the NLLS fits as well. Yet, the deviations are somewhat larger compared to the CRB-based network. 
Further, analyzing the interquartile range as well as the overall range of estimates, we find that the CRB-based network has variations comparable to, and for many samples slightly smaller than the spread of NLLS estimates. In comparison, the spread of the MSE-based network estimates is larger for most samples. 

\subsection{In vivo data}
\begin{figure}[!ht]
\centering
\vspace{-1em}
\includegraphics[scale=0.9]{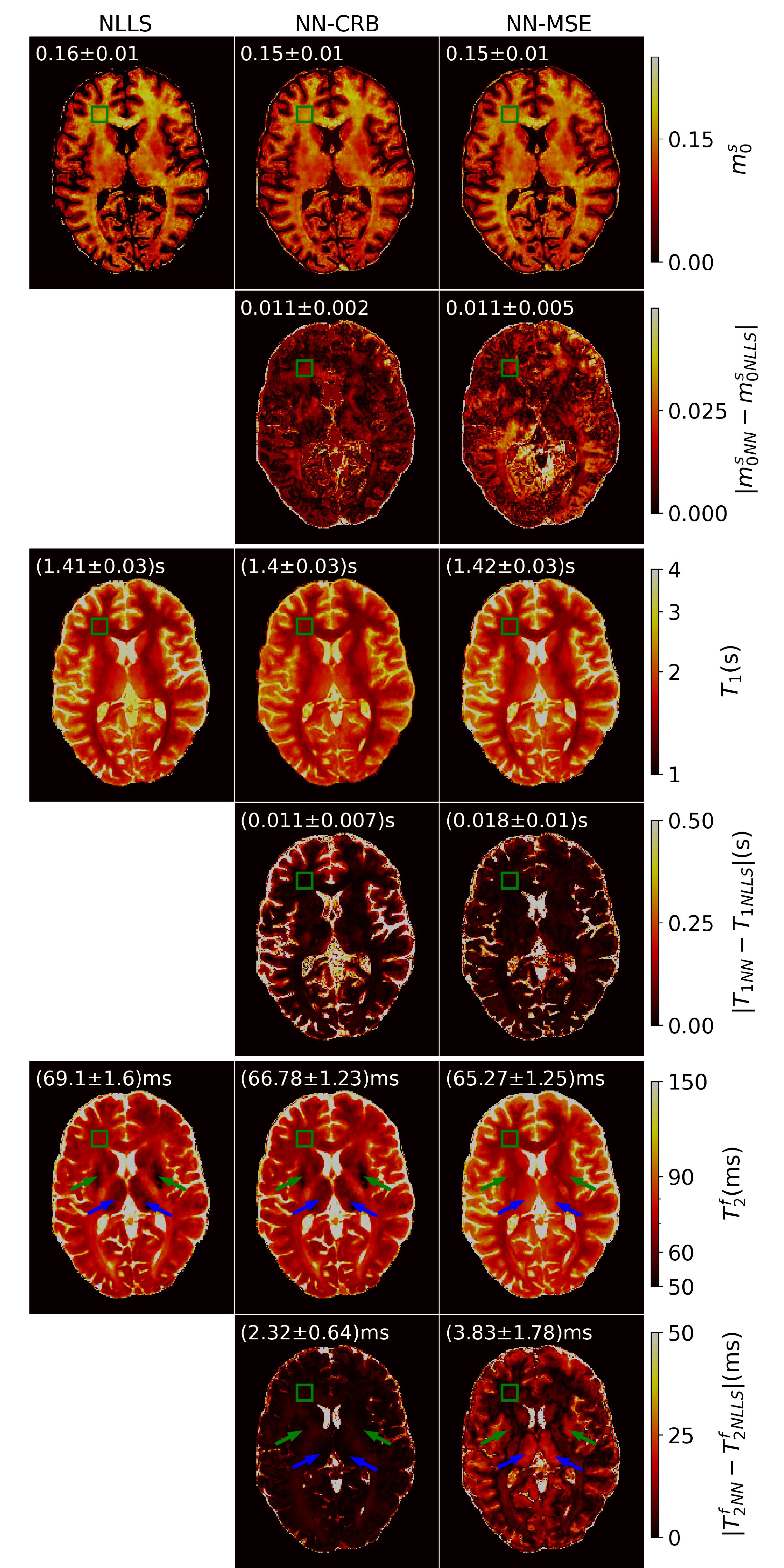}
\caption{A transversal slice through 3D in vivo maps of $m_0^s$, $T_1$ and $T_2^f$, estimated with non-linear least square (NLLS) fitting, a neural network trained the CRB-loss (NN-CRB) and with the MSE-loss (NN-MSE) respectively. The biggest deviations are observed in $T_2^f$ between MSE-based network estimates and NLLS estimates, which we consider the gold standard. The green arrows point to the globus pallidus and the blue arrows point to the thalamus. The MSE-based network does not capture the short $T_2^f$ relaxation times resulting form the iron deposition in those regions. The green rectangle indicates a frontal white matter region of interest (ROI). The mean and standard deviation of the estimates in this ROI can be found in the top left corner of each subfigure.} \label{fig:invivo_nlls}
\end{figure}

The in vivo data paints largely the same picture: We find overall good agreement between both networks and the NLLS fits. For $m_0^s$, the relative deviations between the CRB-based network and NLLS is approximately 10\% and the deviations for the MSE-based network are slightly larger. 
For the $T_1$ estimations, the MSE-based network performs slightly better, which is in line with our finding that the MSE training puts more emphasis on $T_1$. 
The biggest difference is observed in the estimates of $T_2^f$, where the CRB-based network performs substantially better compared to the MSE-based network: 
In the globus pallidus(green arrow in Fig. \ref{fig:invivo_nlls}) and the thalamus(blue arrow in Fig. \ref{fig:invivo_nlls}), the NLLS and CRB-based network estimations show short $T_2^f$ relaxation times as a result of iron deposition\cite{peran2009volume,walsh2014longitudinal}. The MSE-based network is not able to capture this signal variation. 
These findings are also in line with our conceptual and numerical analysis of the networks, which suggested that the MSE-based network performs particularly poorly in $T_2^f$.

%%%%%%%%%%%%%%%%%%%%%%%%%%%%%%%%%%%% Discussion %%%%%%%%%%%%%%%%%%%%%%%%%%%%%%%%%%%%%%%%%%%%%%%%
% What applications is it good for (beneficial for all, in particular complex and high dimensional models

\section{Discussion}
\label{sec:discussion}
As neural networks are increasingly being used to fit biophysical models to MRI data, the need for tailored methods becomes more apparent. Here, we took on the task of finding a training loss that delivers robustness even in heterogeneous parameter spaces. 
We found that off-the-shelf loss functions like the mean squared error over-emphasize estimates that have large values or naturally have a large error, e.g., because the parameter, at this particular value, is not well encoded by the pulse sequence.
The precision with which a parameter is encoded is characterized by the Cram\'er-Rao bound (CRB) and we demonstrated in this paper that normalizing the squared error of each estimate by respective CRB balances the individual contributions to the training loss. 

This normalization of the squared error with the CRB is not entirely heuristic, but rather follows some theoretical concepts: first, it makes the loss of each parameter dimensionless, which allows for adding up the loss of multiple parameters. 
Second, it normalizes the loss by the one of a maximally efficient unbiased estimator, which provides an absolute evaluation metric for a network; and the networks we trained with the CRB-loss indeed converged approximately to the one of a maximally efficient unbiased estimator.  

In order to confirm that our network indeed approximates this ideal condition, we analyzed the bias and the variance of the estimates. We found that the bias of the network trained with the CRB-loss is indeed small (Figs. \ref{fig:T2heatmap}, \ref{fig:validation_bias}, \ref{fig:noise_level}-\ref{fig:phantom}) and that the noise variance closely resembles the CRB (Figs. \ref{fig:T2heatmap} and \ref{fig:validation_variance}), which indicates that the CRB-based network indeed approximates a maximally-efficient unbiased estimator. 
Further, we found that this approximation is much better compared to a network that was trained with the MSE-loss. 

Here, we tested the CRB-loss function at the example of an 8-parameter magnetization transfer model \cite{Asslander2019ismrm}. 
The theoretical foundation of the proposed loss function gives us reason to believe that it results in superior performance for any model, but the necessity for a well-balanced loss function certainly grows with the heterogeneity of the parameter space or, more precisely, with increasing variations of the Cram\'er-Rao bound between different parameters and/or throughout the parameter space. 

We calculated here the CRB assuming independent and identically distributed Gaussian noise, an assumption that is also implicitly baked into the MSE loss. 
In order to approximately fulfill this assumption, we reconstructed images into a low rank space spanned by the SVD of the training data \cite{tamir2017t2,asslander2018low} and used a combination of parallel imaging \cite{Sodickson1997,Pruessmann2001} and locally low rank flavored compressed sensing \cite{Lustig2008,Trzasko2011a,tamir2017t2} to virtually remove the undersampling artifacts. 
This allows us to train the network with additive Gaussian noise rather than relying on a heuristic noise statistics that emulate undersampling artifacts\cite{virtue2017better,virtue2018learning}. 

%%%%%%%%%%%%%%%%%%%%%% Conclusion %%%%%%%%%%%%%%%%%%%%%%%%%%%%%%%%%%%%%%%%%%%%%%%%%%%%%%%%%%
To conclude, we have introduced a theoretically well-founded loss function for deep-learning-based method of parameter estimation in quantitative MRI, and demonstrated its superior performance when compared to the commonly used MSE loss function.

\bibliography{reference}  %%% Uncomment this line and comment out the ``thebibliography'' section below to use the external .bib file (using bibtex) .

%%% Uncomment this section and comment out the \bibliography{references} line above to use inline references.
% \begin{thebibliography}{1}

% 	\bibitem{kour2014real}
% 	George Kour and Raid Saabne.
% 	\newblock Real-time segmentation of on-line handwritten arabic script.
% 	\newblock In {\em Frontiers in Handwriting Recognition (ICFHR), 2014 14th
% 			International Conference on}, pages 417--422. IEEE, 2014.

% 	\bibitem{kour2014fast}
% 	George Kour and Raid Saabne.
% 	\newblock Fast classification of handwritten on-line arabic characters.
% 	\newblock In {\em Soft Computing and Pattern Recognition (SoCPaR), 2014 6th
% 			International Conference of}, pages 312--318. IEEE, 2014.

% 	\bibitem{hadash2018estimate}
% 	Guy Hadash, Einat Kermany, Boaz Carmeli, Ofer Lavi, George Kour, and Alon
% 	Jacovi.
% 	\newblock Estimate and replace: A novel approach to integrating deep neural
% 	networks with existing applications.
% 	\newblock {\em arXiv preprint arXiv:1804.09028}, 2018.

% \end{thebibliography}

\newpage
\appendix

\begin{center}
\LARGE{Supplementary Material}
\end{center}
\vspace{1cm}
%heatmap m0s
\begin{figure}[htbp]
\centering
\vspace{-1em}
\includegraphics[scale=0.35]{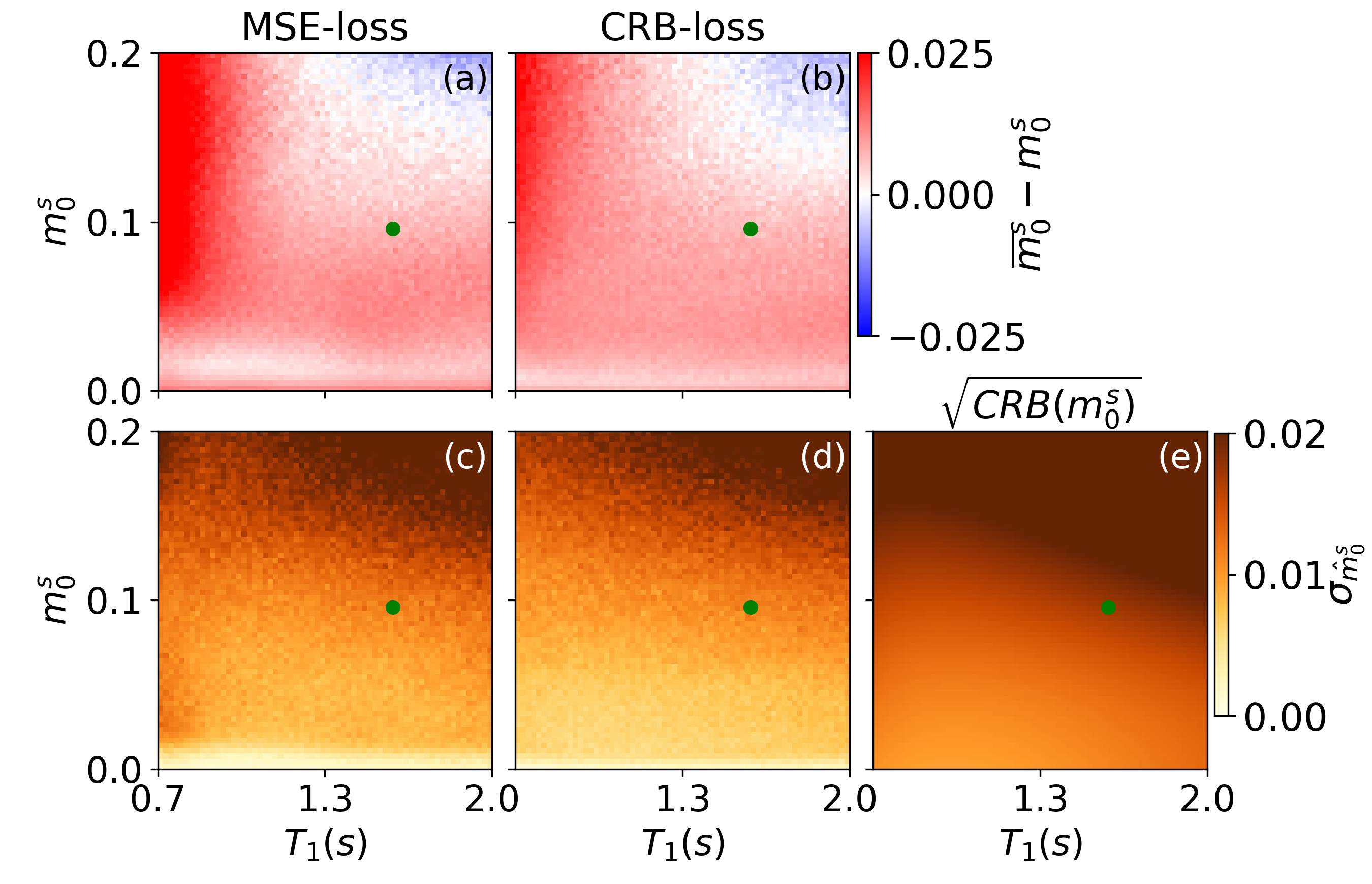}
\caption{Bias (a,b) and standard deviation (c,d) of $\hat{m}_0^s$, estimated with the network trained with the MSE-loss and CRB-loss, respectively. The latter is compared to the square root of the Cram\'er-Rao bound (e), which provides a theoretical limit for an unbiased estimator. The green dots indicate the mean values of the corresponded parameters in the training dataset.
The maps were generated with the test dataset \#2.} 
\label{fig:m0sheatmap}
\end{figure}

%heatmap T1
\begin{figure}[htbp]
\centering
\vspace{-1em}

\includegraphics[scale=0.35]{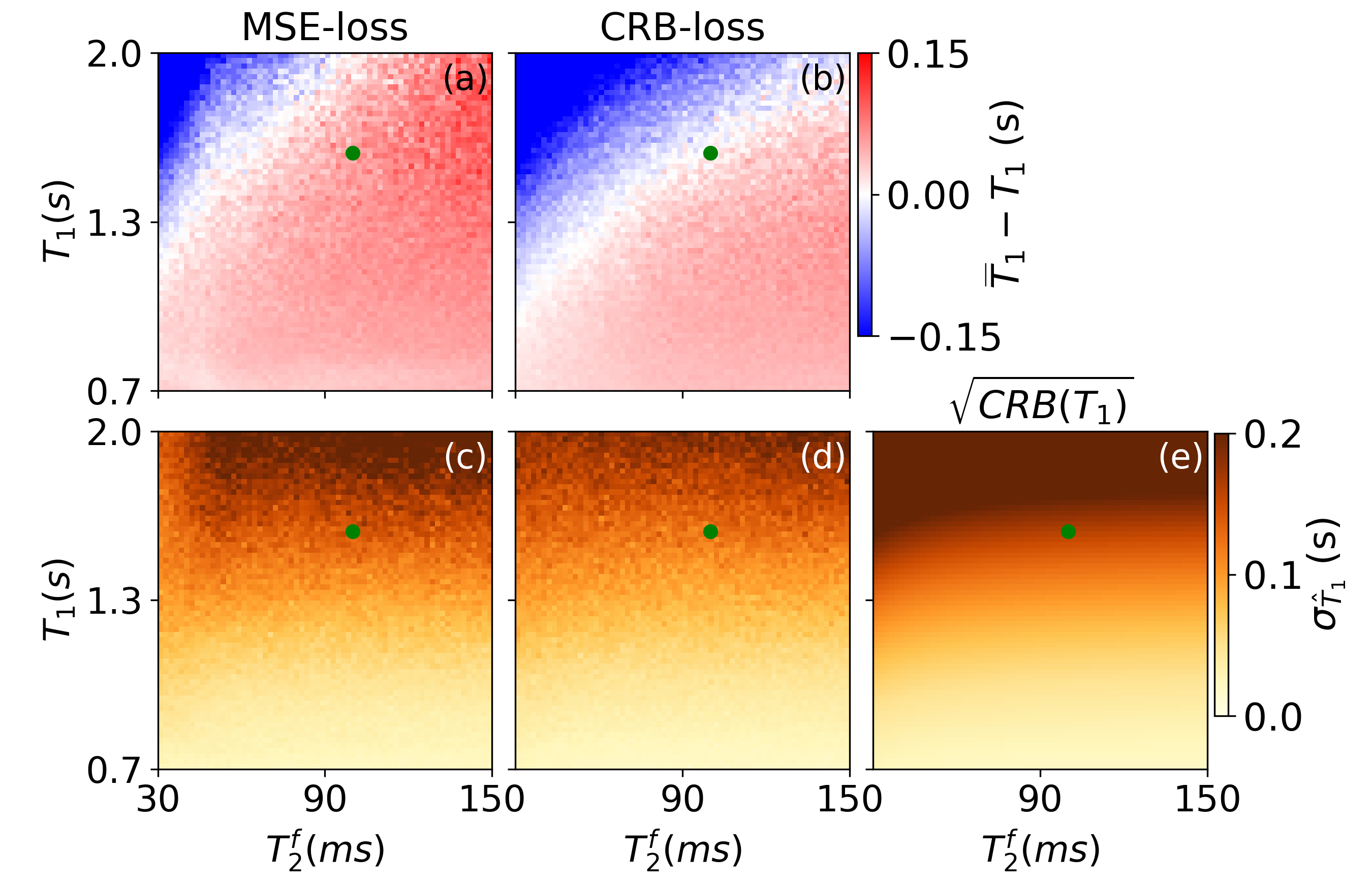}
\caption{Bias (a,b) and standard deviation (c,d) of $\hat{T}_1$, estimated with the network trained with the MSE-loss and CRB-loss, respectively. The latter is compared to the square root of the Cram\'er-Rao bound (e), which provides a theoretical limit for an unbiased estimator.The green dots indicate the mean values of the corresponded parameters in the training dataset. 
The maps were generated with the test dataset \#2.} \label{fig:T1heatmap}
\end{figure}

When using the network trained with the CRB-loss, the bias of the $T_1$ estimation (Supplementary Fig.~\ref{fig:T1heatmap}b) with the CRB-based network is slightly larger (0.11 on average of absolute bias) than that of the MSE-based network (Fig.~\ref{fig:T1heatmap}a; 0.07 on average), which is in line with the hypothesis that the MSE loss puts more emphasis on $T_1$. The bias of the $m_0^s$ estimation with the CRB-based network in a slice spanned by $m_0^s$ and $T_1$ (Fig.~\ref{fig:m0sheatmap}b) is slightly smaller (0.0053 on average of absolute bias) than that of the MSE-based network (Fig.~\ref{fig:m0sheatmap}a; 0.0065 on average). Overall, the two networks have similar performance in estimating $m_0^s$ and $T_1$, while we do observe a substantial difference in the performance in estimating $T_2^f$.

\end{document}